\documentclass[smallextended]{svjour3}       
\smartqed  
\usepackage{graphicx}
\usepackage{tabularx}
\usepackage{array}
\usepackage[round]{natbib}
\usepackage{verbatim}

\usepackage{subcaption}
\usepackage{graphicx}
\usepackage{xcolor,colortbl}
\usepackage{caption}

\usepackage{listings}
\usepackage{url}
\usepackage{paralist}
\usepackage{multirow}

\usepackage{color}
\usepackage{hyperref}
\hypersetup{
    colorlinks=true, 
    linktoc=all,     
    linkcolor=black,  
    citecolor=black,
    filecolor=blue,
    urlcolor=blue
}

\renewcommand*\contentsname{This Summary will be removed later}

\renewcommand{\cite}{\citep}

\emergencystretch=\maxdimen
\begin{document}

\title{Meta-QSAR: a large-scale application of meta-learning to drug design and discovery}



\author{Ivan Olier         \and
        Noureddin Sadawi \and 
        G. Richard Bickerton \and 
        Joaquin Vanschoren \and 
        Crina Grosan \and
        Larisa Soldatova \and 
        Ross D. King 
}


\institute{
              I. Olier \at
              Manchester Metropolitan University and University of Manchester
              \email{i.olier@mmu.ac.uk}           
        \and
              N. Sadawi \at
              Imperial College London and Brunel University London
              \email{n.sadawi@imperial.ac.uk}
        \and
              G. R. Bickerton \at
              Dundee University
              \email{g.r.bickerton@dundee.ac.uk}
              \at 
              exscientia Ltd
              \email{rbickerton@exscientia.co.uk}
        \and
              J. Vanschoren \at
              Eindhoven University of Technology
              \email{j.vanschoren@tue.nl}     
        \and
              C. Grosan \at
              Brunel University London
              \email{crina.grosan@brunel.ac.uk}
        \and
              L. Soldatova \at
              Brunel University London
              \email{larisa.soldatova@brunel.ac.uk}  
        \and
              R. D. King \at
              University of Manchester
              \email{r.king@manchester.ac.uk}  
}

\date{Received: date / Accepted: date}

\maketitle

\begin{abstract}
We investigate the learning of quantitative structure activity relationships (QSARs) as a case-study of meta-learning. This application area is of the highest societal importance, as it is a key step in the development of new medicines. The standard QSAR learning problem is: given a target (usually a protein) and a set of chemical compounds (small molecules) with associated bioactivities (e.g. inhibition of the target), learn a predictive mapping from molecular representation to activity. Although almost every type of machine learning method has been applied to QSAR learning there is no agreed single best way of learning QSARs, and therefore the problem area is well-suited to meta-learning. 
 
We first carried out the most comprehensive ever comparison of machine learning methods for QSAR learning: 18 regression methods, 6 molecular representations, applied to more than 2,700 QSAR problems. 
(These results have been made publicly available on OpenML and represent a valuable resource for testing novel meta-learning methods.) 
We then investigated the utility of algorithm selection for QSAR problems. We found that this meta-learning approach significantly outperformed the best individual QSAR learning method (random forests using a molecular fingerprint representation). We conclude that meta-learning outperforms base-learning methods for QSAR learning, and as this investigation is one of the most extensive ever comparisons of base and meta-learning methods ever made, it provides evidence for the general effectiveness of meta-learning over base-learning. 
\keywords{Meta-learning \and Algorithm selection \and Drug Discovery \and QSAR}
\end{abstract}

\section{Introduction}
\label{sec:intro}
The standard approach to predicting how active a chemical compound will be against a given target (usually a protein that needs to be inhibited) in the development of new medicines is to use machine learning models. Currently, there is no agreed single best learning algorithm to do this. In this paper we investigate the utility of meta-learning to address this problem. We aim to discover and exploit relationships between machine learning algorithms, measurable properties of the input data, and the empirical performance of learning algorithms, to infer the best models to predict the activity of chemical compounds on a given target. 

\subsection{Quantitative Structure Activity Relationship (QSAR) Learning}
\label{qsar-learning}
Drug development is one of the most important applications of science, as it is an essential step in the treatment of almost all diseases. Developing a new drug is however slow and expensive. The average cost to bring a new drug to market is $>$ 2.5 billion US dollars (Tufts, 2014), which means that tropical diseases such as malaria, schistosomiasis, Chagas' disease, etc., which kill millions of people and infect hundreds of millions of others are `neglected' (Ioset \& Chang, 2011; Leslie, 2011) and that 
`orphan' diseases (i.e. those with few sufferers) remain untreatable (Braun et al, 2010). More generally, the pharmaceutical industry is struggling to cope with spiralling drug discovery and development costs (Pammolli et al, 2011). Drug development is also slow, generally taking more than 10 years. This means that there is strong pressure to speed up development, both to save lives and reduce costs. A successful drug can earn billions of dollars a year, and as patent protection is time-limited, even one extra week of patent protection can be of great financial significance.

A key step in drug development is learning Quantitative Structure Activity Relationships (QSARs) (Martin, 2010),\cite{Cherkasov:2014go,Cumming:2013eu}.  These are functions that predict a compound's bioactivity from its structure. The standard QSAR learning problem is: given a target (usually a protein) and a set of chemical compounds (small molecules) with associated bioactivities (e.g. inhibiting the target), learn a predictive mapping from molecular representation to activity. 

Although almost every form of statistical and machine learning method has been applied to learning QSARs, there is no agreed single best way of learning QSARs. Therefore an important motivation for this work is to better understand the performance characteristics of the main (baseline) machine learning methods currently used in QSAR learning. This knowledge will feed into a better understanding of the performance characteristics of these algorithms, and will enable QSAR practitioners to improve there predictions. 

The central motivation for this work is to better understand meta-learning through a case-study in the very important real-world application area of QSAR learning. This application area is an excellent test-bed for the development of meta-learning methodologies. The importance of the subject area means that there are now thousands of publicly available QSAR datasets, all with the same basic structure. Few machine learning application areas have so many datasets - enabling statistical confidence in meta-learning results. 
In investigating meta-learning we have focused on algorithm selection as this is the simplest form of meta-learning, and its use fits in with our desire to better understand the baseline-learning methods. 

A final motivation for the work is to improve the predictive performance of QSAR learning through use of meta-learning.  Our hope is that improved predictive performance will feed into faster and cheaper drug development.

To enable others to build on our base-learning and meta-learning work we have placed all our results in OpenML.

\subsection{Meta-Learning: Algorithm Selection}
\label{meta-algo}
Meta-learning has been used extensively to select the most appropriate learning algorithm on a given dataset. In this section, we first sketch a general framework for algorithm selection, and then provide an overview of prior approaches and the state-of-the-art in selecting algorithms using meta-learning.

\subsubsection{Algorithm Selection Framework}
\label{framework}

\begin{figure}
\center
\includegraphics[width=0.7\textwidth]{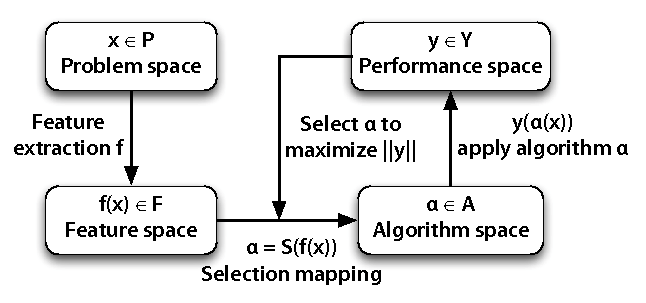}
\caption{Rice's framework for algorithm selection. Adapted from \protect \cite{Rice76,Smith2008}.}
\label{rice}
\end{figure}

The algorithm selection framework contains four main components:
\emph{First}, we construct the problem space $P$, in our case the space of all QSAR datasets. Each dataset expresses the properties and activity of a limited set of molecular compounds (drugs) on a specific target protein. In this paper, we consider 2,764 QSAR datasets, described in more detail in Section~\ref{baseline-qsar-datasets}.
\emph{Second}, we describe each QSAR dataset in $P$ with a set of measurable characteristics (meta-features), yielding the feature space $F$. In this paper we include two types of meta-features: those that describe the QSAR data itself (e.g. the number of data points), and those that describe properties of the target protein (e.g. hydrophobicity). We expect that these properties will affect the interplay of different QSAR features, and hence the choice of learning algorithm. The full set of meta-features used in this paper is described in Section~\ref{sec:metafeatures}.

\emph{Third}, the algorithm space $A$ is created by the set of all candidate base-level learning algorithms, in our case a set of 18 regression algorithms combined with several preprocessing steps. These are described in Section~\ref{baseline-algorithms}.

\emph{Finally}, the performance space $Y$ represents the empirically measured performance, e.g. root mean squared error (RMSE) \cite{Witten:2005:DMP:1205860} of each algorithm $A$ on each of the QSAR datasets in $P$.

In the current state-of-the-art, there exists a wide variety of algorithm selection algorithms. If only a single algorithm should be run, we can train a classification model that makes exactly that prediction \cite{Pfahringer00, Guerri04}. We can also use a regression algorithm to predict the performance of each algorithm \cite{Xu08}, build a ranking of promising algorithms \cite{LeiteBV12}, or use cost-sensitive techniques which allow us to optimize the loss we really care about in the end \cite{Bischl12, Xu12}.

Our task is: for any given QSAR problem $x \in P$, select the best combination of QSAR and molecular representation $ a \in A$ that maximizes a predefined performance measure $y \in Y$. In this paper, we investigate two meta-learning approaches: 1) classification problem: the aim is to learn a model that captures the relationship between the properties of the QSAR datasets, or meta-data, and the performance of the regression algorithms. This model can then be used to predict the most suitable algorithm for a new dataset. 2) ranking problem: the aim is to fit a model that ranks the QSAR combinations by their predicted performances.

\subsubsection{Previous Work on Algorithm Selection using Meta-Learning}
\label{previous-work}
\paragraph{Meta-features.}
In the meta-learning literature much effort has been devoted to the development of meta-features that effectively describe the characteristics of the data. These should have discriminative power, meaning that they should be able to distinguish between base-learners in terms of their performance, and have a low computational complexity - preferably lower than $O(n \log n)$ \cite{Pfahringer00}. Meta-features are typically categorised as one of the following: simple (e.g. number of data points, number of features), statistical (e.g. mean standard deviation of attributes, mean kurtosis of attributes, mean skewness of attributes), or information theoretic (e.g. mean entropy of the features, noise-signal ratio). See \cite{Bickel:2008:MLH:1390156.1390164,Kalousis:2002p336,Vanschoren:2010phd} for an extensive description of meta-features. A subset of these may be used for regression, and some measures are specifically defined for regression targets \cite{Soares:2004p700}. Other meta-features can be trivially adapted to the regression data. First, landmarking \cite{Pfahringer00} works by training and evaluating sets of simple, fast algorithms on the datasets (e.g. a decision stump instead of a full decision tree), and using their performance (e.g. RMSE) as meta-features for the dataset. An analysis of landmarkers for regression problems can be found in \citet{Ler:2005p1680}.

Another approach is to use model-based characteristics \cite{Peng:2002p694}, obtained by building fast, interpretable models, e.g. decision trees, and then extracting properties of those models, such as the width, the depth and the number of leaves in the tree, and statistical properties (min, max, mean, stdev) of the distribution of nodes in each level of the tree, branch lengths, or occurrences of features in the splitting tests in the nodes. Recent research on finding interesting ways to measure data characteristics includes instance-level complexity \cite{SmithMG14}, measures for unsupervised learning \cite{LeeG11}, and discretized meta-features \cite{LeeG08a}. 

Meta-learning has also been successfully applied in stream mining \cite{RijnHPV15,RijnHPV14} and time series analysis \cite{Prudencio:2004p6308}, each time requiring novel sets of meta-features.

\paragraph{Selecting algorithms.}
In meta-learning, algorithm selection is traditionally seen as a learning problem: train a meta-learner that predicts the best algorithm(s) given a set of meta-features describing the data. In the setting of selecting a best single algorithm, experiments on artificial datasets showed that there is no single best meta-learner, but that decision tree-like algorithms (e.g. C5.0boost) seem to have an edge, especially when used in combination with landmarkers \cite{Bensusan:2000p557,Pfahringer00}. Further experiments performed on real-world data corroborated these results, although they also show that most meta-learners are very sensitive to the exact combination of meta-features used \cite{Kopf:2000p5809}.
					
In the setting of recommending a subset of algorithms it was shown that, when using statistical and information-theoretical meta-features, boosted decision trees obtained best results \cite{Kalousis:2002p336,Kalousis:2001p314}. Relational case-based reasoning has also been successfully applied \cite{Lindner:1999p9438,Hilario:2001p001}, which allows to include algorithm properties independent of the dataset and histogram representations of dataset attribute properties.

Most relevant for this paper is the work by Amasyali and Ersoy \cite{FatihAmasyali:2009}, which uses around 200 meta-features to select the best regression algorithm for a range of artificial, benchmarking, and drug discovery datasets. The reported correlations between meta-features and algorithm performances were typically above 0.9 on artificial and benchmarking datasets, but much worse (below 0.8) on the drug discovery datasets. Feature selection was found to be important to improve meta-learning performance.

\paragraph{Ranking algorithms}
Another approach is to build a ranking of algorithms, listing which algorithms to try first. Several techniques use k-nearest neighbors \cite{Brazdil:2003p697,dosSantos:2004p6355}, and compute the average rank (or success rate ratio's or significant wins) over all similar prior datasets \cite{Soares:2000p720,Brazdil:2000p688}. Other approaches directly estimate the performances of all algorithms \cite{Bensusan:2001p344}, or use predictive clustering trees \cite{Todorovski:2002p6085}.

Better results where obtained by \textit{subsampling landmarkers}, i.e. running all candidate algorithms on several small samples of the new data \cite{Furnkranz:2001p1278}. Meta-learning on data samples (MDS) \cite{Leite:2005p725,Leite:2007p6022} builds on this idea by first determining the complete learning curves of a number of learning algorithms on several different datasets. Then, for a new dataset, progressive subsampling is done up to a certain point, creating a partial learning curve, which is then matched to the nearest complete learning curve for each algorithm in order to predict their final performances on the entire new dataset.

Another approach is to sequentially evaluate a few algorithms on the (complete) new dataset and learn from these results. \textit{Active testing} \cite{LeiteBV12} proceeds in a tournament-style fashion: in each round it selects and tests the algorithm that is most likely to outperform the current best algorithm, based on a history of prior duels between both algorithms on similar datasets. Each new test will contribute information to a better estimate of dataset similarity, and thus help to better predict which algorithms are most promising on the new dataset. Large-scale experiments show that active testing outperforms previous approaches, and yields an algorithm whose performance is very close to the optimum, after relatively few tests. More recent work aims to speed up active testing by combining it with learning curves \cite{RijnABV15}, so that candidates algorithms only need to be trained on a smaller sample of the data. It also uses a multi-objective criterion called AR3 \cite{AbdulrahmanB14} that trades off runtime and accuracy so that fast but reasonably accurate candidates are evaluated first. Experimental results show that this method converges extremely fast to an acceptable solution.

Finally, algorithms can also be ranked using collaborative filtering \cite{bardenet:in2p3-00907381,misir:hal-00922840,SmithMGM14}. In this approach, previous algorithm evaluations are used as `ratings' for a given dataset. For a new dataset, algorithms which would likely perform well (give a high rating) are selected based on collaborative filtering models (e.g. using matrix decompositions).

\paragraph{Model-based optimization}
Model-based optimization \cite{HutHooLey11} aims to select the best algorithm and/or best hyperparameter settings for a given dataset by sequentially evaluating them on the full dataset. It learns from prior experiments by building a surrogate model that predicts which algorithms and parameters are likely to perform well. An approach that has proven to work well in practice is Bayesian Optimization \cite{brochu2010tutorial}, which builds a surrogate model (e.g. using Gaussian Processes or Random Forests) to predict the expected performance of all candidate configurations, as well as the uncertainty of that prediction. In order to select the next candidate to evaluate, an acquisition function is used that trades off exploitation (choosing candidates in regions known to perform well) versus exploration (trying candidates in a relatively unexplored regions). Bayesian Optimization is used in Auto-WEKA \cite{ThoHutHooLey13} and Auto-sklearn \cite{Feurer2015}, which search for the optimal algorithms and hyperparameters across the \textit{WEKA}~\cite{Hall:2009:WDM:1656274.1656278} and \textit{scikit-learn}~\cite{scikit-learn} environments, respectively. Given that this technique is computationally very expensive, recent research has tried to include meta-learning to find a good solution faster. One approach is to find a good set of initial candidate configurations by using meta-learning \cite{Feurer2015}: based on meta-features, one can find the most similar datasets and use the optimal algorithms and parameter settings for these datasets as the initial candidates to evaluate. In effect, this provides a `warm start' which yields better results faster.

\subsection{Meta-QSAR Learning}\label{meta-qsar-learning}
Almost every form of statistical and machine learning method has been applied to learning QSARs: linear regression, decision trees, neural networks, nearest-neighbour methods, support vector machines, Bayesian networks, relational learning, etc.  These methods differ mainly in their \textit{a priori} assumptions they make about the learning task.  We focus on regression algorithms as this is how QSAR problems are normally cast.  

For Meta-QSAR learning the input data are datasets of compound activity (one for each target protein), different representations of the structures of the compounds, and we aim to learn to predict how well different learning algorithms perform, and to exploit these predictions to improve QSAR predictions. We expect meta-learning to be successful for QSAR because although all the datasets have the same overall structure, they differ in the numbers of data points (tested chemical compounds), in the range and occurrence of features (compound descriptors), and in the type of chemical/biochemical mechanism that causes the bioactivity.  These differences indicate that different machine learning methods are to be used for different kinds of QSAR data.

We first applied meta-learning to predict the machine learning algorithm that is expected to perform best on a given QSAR dataset. This is known as the algorithm selection problem, and can be expressed formally using Rice's framework for algorithm selection \cite{Rice76} as illustrated in Figure~\ref{rice}.
We then applied multi-task learning to first test whether it can improve on standard QSAR learning through the exploitation of evolutionary related targets, and whether multi-task learning can further be improved by incorporating the evolutionary distance of targets.\\

\subsection{Paper Outline}
The remainder of this paper is organized as follows. In Section~\ref{sec:baseline}, we report our baseline experiments investigating the effectiveness of a large number of regression algorithms on thousands of QSAR datasets, using different data representations. In Section~\ref{sec:metafeatures} we describe a novel set of QSAR-specific meta-features to inform our meta-learning approach. In Section~\ref{sec:meta-algo} we investigate the utility of meta-learning for selecting the best algorithm for learning QSARs. Finally,  Section~\ref{sec:discussion} presents a discussion of our results and future work.

\section{Baseline QSAR Learning} \label{sec:baseline}
We first performed experiments with a set of baseline regression algorithms to investigate their effectiveness on QSAR problems. Learning a QSAR model consists of fitting a regression model to a dataset which has as instances the chemical compounds, as input variables the chemical compound descriptors, and as numeric response variable (output) the associated bioactivities.  

\subsection{Baseline QSAR Learning Algorithms}
\label{baseline-algorithms}
For our baseline QSAR methods we selected 18 regression algorithms, including linear regression, support vector machines, artificial neural networks, regression trees, and random forests.
Table \ref{table:baseline-qsar-algos} lists all the algorithms used and their respective parameter settings. Within the scope of this study, we do not optimize the parameter settings on every dataset, but instead chose values that are likely to perform well on most QSAR datasets. This list includes the most commonly used QSAR methods in the literature.

With the exception of one of the neural networks implementations, for which we used the H2O R package\footnote{\url{https://cran.r-project.org/web/packages/h2o/index.html}}, all of the algorithms were implemented using the MLR R package for machine learning\footnote{\url{https://cran.r-project.org/web/packages/mlr/index.html}}.

\begin{center}
\begin{table}

    \begin{tabularx}{\columnwidth}{ | l | >{\raggedright\arraybackslash} X | >{\raggedright\arraybackslash} X | }
    \hline
\textbf{Short name} & \textbf{Name} & \textbf{Parameter settings} \\ \hline
ctree & Conditional trees & min\_split=20, min\_bucket=7\\ \hline
rtree & Regression trees & min\_split=20, min\_bucket=7\\ \hline
cforest & Random forest (with conditional trees) & n\_trees=500, min\_split=20, min\_bucket=7\\ \hline
rforest & Random forest & n\_trees=500, min\_split=20, min\_bucket=7\\ \hline
gbm & Generalized boosted regression & n\_trees=100, depth=1, CV=no, min\_obs\_node=10\\ \hline
fnn & k-Nearest neighbor & k=1\\ \hline
earth & Adaptive regression splines (earth) & (as default)\\ \hline
glmnet &  Regularized GLM & (as default)\\ \hline
ridge & Penalized ridge regression & (as default)\\ \hline
lm & Multiple linear regression & (as default)\\ \hline
pcr & Principal component regression & (as default)\\ \hline
plsr & Partial least squares & (as default)\\ \hline
rsm & Response surface regression & (as default)\\ \hline
rvm & Relevance vector machine & Kernel=RBF, nu=0.2, epsilon=0.1\\ \hline
ksvm & Support vector machines & Kernel=RBF, nu=0.2, epsilon=0.1\\ \hline
ksvmfp & Support vector machines with Tanimoto kernel & Kernel=Tanimoto\\ \hline
nnet &  Neural networks & size=3\\ \hline
nneth2o & Neural networks using H2O library & layers=2, size layer 1 = 0.333* n\_inputs, layer 2 = 0.667*n\_inputs\\ \hline
\end{tabularx}
\caption{List of baseline QSAR algorithms. Abbreviations: n\_trees: number of trees; min\_split: minimum node size allowed for splitting; min\_bucket: minimum size of the bucket. k: number of neighbours; depth: search depth; CV: cross-validation; min\_obs\_node: minimum number of observations per node; RBF: radial basis function with nu (spread) and epsilon (scale) parameters; size: number of neurons in the hidden layer; n\_inputs: length of the input vector.}
\label{table:baseline-qsar-algos}
\end{table}
\end{center}

\subsection{Baseline QSAR Datasets}
\label{baseline-qsar-datasets}
For many years, QSAR research was held back by a lack of openly available datasets. This situation has been transformed by a number of developments.  The most important of these is the open availability of the ChEMBL database\footnote{\url{https://www.ebi.ac.uk/chembl/}}, a medicinal chemistry database managed by the European Bioinformatics Institute (EBI).  It is abstracted and curated from the scientific literature, and covers a significant fraction of the medicinal chemistry corpus.  
The data consist of information on the drug targets (mainly proteins from a broad set of target families, e.g. kinases), the structures of the tested compounds (from which different chemoinformatic representations may be calculated), and the bioactivities of the compounds on their targets, such as binding constants, pharmacology, and toxicity.  
The key advantages of using ChEMBL for Meta-QSAR are: (a) it covers a very large number of targets, (b) the diversity of the chemical space investigated,  and (c) the high quality of the interaction data.  Its main weakness is that for any single target, interaction data on only a relatively small number of compounds are given.  

We extracted 2,764 targets from ChEMBL with a diverse number of chemical compounds, ranging from 10 to about 6,000, each target resulting in a dataset with as many examples as compounds. The target (output) variable contains the associated bioactivities.  Bioactivity data were selected on the basis that the target type is a protein, thereby excluding other potential targets such as cell-based and \textit{in vivo} assays, and the activity type is from a defined list of potency/affinity endpoints (IC50, EC50, Ki, Kd and their equivalents). In the small proportion of cases where multiple activities have been reported for a particular compound-target pair, a consensus value was selected as the median of those activities falling in the modal log unit. The simplified molecular-input line-entry system (SMILES) representation of the molecules was used to calculate molecular properties such as molecular weight (MW), logarithm of the partition coefficient (LogP), topological polar surface area (TPSA), etc. For this we used Dragon version 6 \cite{Mauri:2006}, which is a commercially available software library that can potentially calculate up to 4,885 molecular descriptors, depending on the availability of 3D structural information of the molecules. A full list is available on Dragon's website\footnote{\url{http://www.talete.mi.it}}. 

As ChEMBL records 2D molecular structures only, we were restricted to estimating a maximum of 1,447 molecular descriptors. We decided to generate datasets using all permitted molecular descriptors as features, and then to extract a subset of 43, which Dragon identifies as basic or constitutional descriptors. We call these representations `allmolprop' and `basicmolprop', respectively. For some of the molecules, Dragon failed to compute some of the descriptors, possibly because of bad or malformed structures, and these were treated as missing values. To avoid favouring QSAR algorithms able to deal with missing values, we decided to impute them, as a preprocessing step, using the median value of the corresponding feature.
 
In addition, we calculated the FCFP4 fingerprint representation using the Pipeline Pilot software from BIOVIA \cite{Rogers:2010cy}. The fingerprint representation is the most commonly used in QSAR learning, whereby the presence or absence of a particular molecular substructure in a molecule (e.g. methyl group, benzine ring) is indicated by a Boolean variable. The FCFP4 fingerprint implementation generates 1024 such Boolean variables.  We call this dataset representation `fpFCFP4'. All of the fpFCFP4 datasets were complete, so a missing value imputation step is not necessary.

In summary, we use 3 types of feature representations and 1 level of preprocessing, thus generating 3 different dataset representations for each of the QSAR problems (targets), see Table \ref{table:Dataset Representations}. This produced in total 8,292 datasets from the 2,764 targets.

\begin{center}
\begin{table}[t]
    \begin{tabularx}{\columnwidth}{ | >{\raggedright\arraybackslash} X | >{\raggedright\arraybackslash} X | >{\raggedright\arraybackslash} X | >{\raggedright\arraybackslash} X |}
    \hline
 & Basic set of descriptors (43) & All descriptors (1447) & FCFP4 fingerprint (1024) \\ \hline
Original dataset  & basicmolprop (not used) & allmolprop (not used) & fpFCFP4 \\ \hline
Missing value imputation & basicmolprop.miss & allmolprop.miss & (no missing values) \\ \hline
\end{tabularx}
\caption{Names of the generated dataset representations.
}
\label{table:Dataset Representations}
\end{table}
\end{center}

\subsection{Baseline QSAR Experiments}
\label{baseline-qsar-exp}
The predictive performance of all the QSAR learning methods on the datasets (base QSAR experiments) was assessed by taking the average root mean squared error (RMSE) with 10-fold cross-validation. 

\begin{figure}
\center
\includegraphics[width=0.8\textwidth]{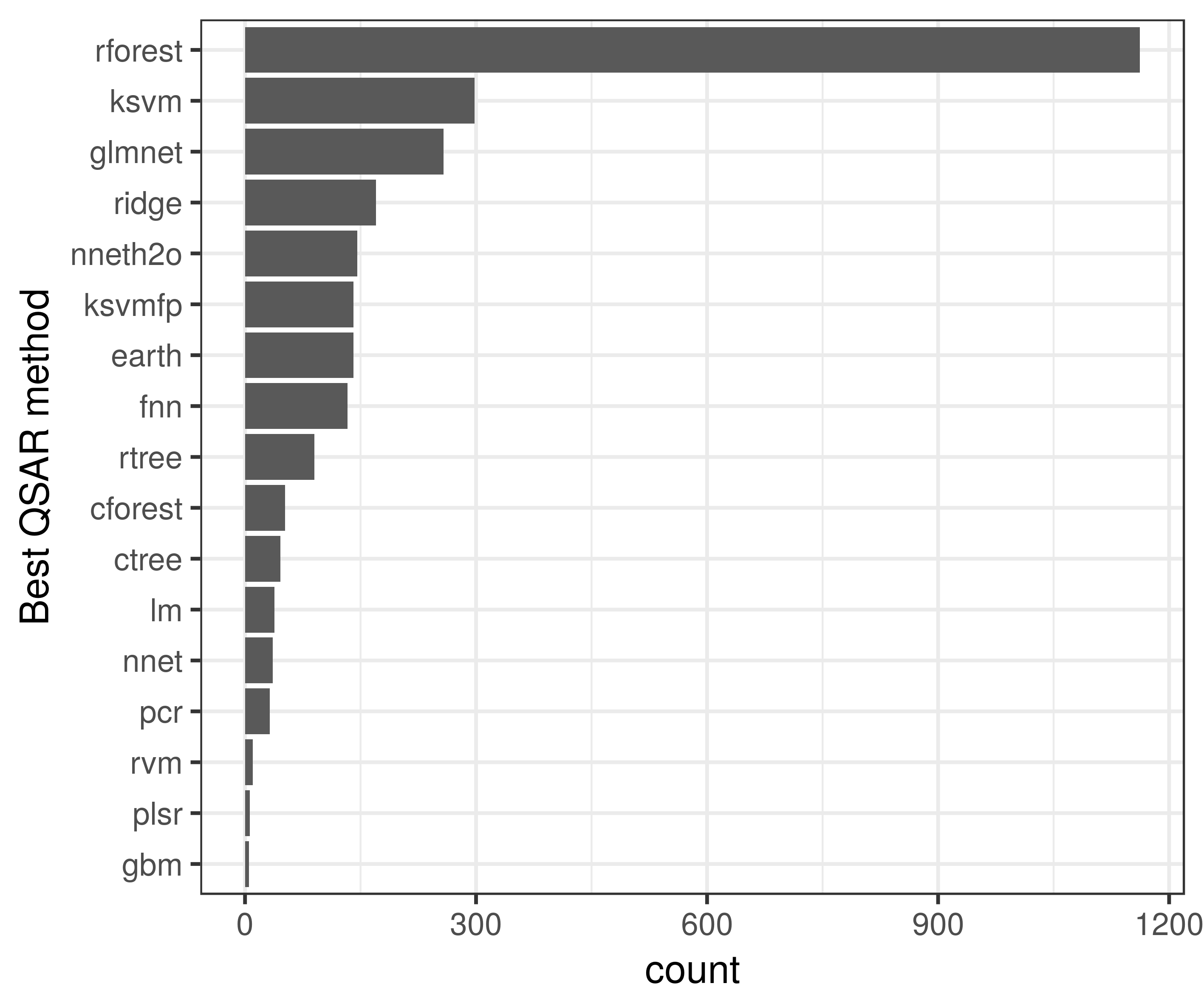}
\caption{Graphical representation of the number of times (target counts) a particular QSAR learning method obtains the best performance (minimum RMSE)}
\label{fig:fig_baseQSARperf_algorithms}
\end{figure}

\begin{figure}
\center
\includegraphics[width=0.7\textwidth]{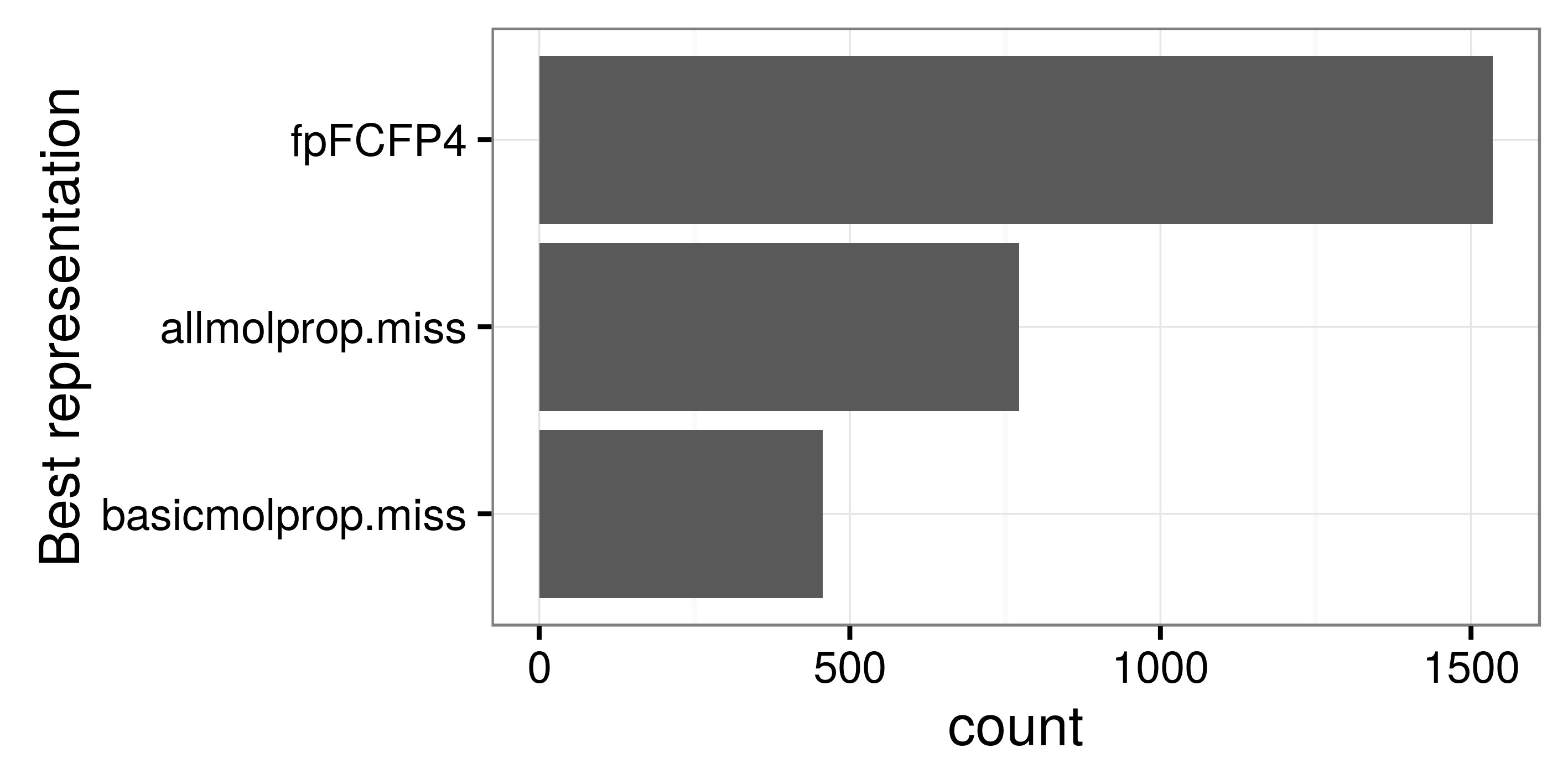}
\caption{Graphical representation of the number of times (target counts) a dataset representation was fitted with the best performer QSAR method (minimum RMSE)}
\label{fig:fig_baseQSARperf_repres}
\end{figure}

We used the parameter settings mentioned in Table \ref{table:baseline-qsar-algos} for all experiments. Figure \ref{fig:fig_baseQSARperf_algorithms} summarizes the overall relative performance  (in frequencies) of the QSAR methods for all dataset representations previously mentioned in Table \ref{table:Dataset Representations}. Results showed that random forest (`rforest') was the best performer in 1,162 targets out of 2,764, followed by SVM (`ksvm'), 298 targets, and GLM-NET (`glmnet'), 258 targets.  In these results, the best performer is the algorithm with the lowest RMSE, even if it wins by a small margin. In terms of dataset representation, it turned out that datasets formed using FCFP4 fingerprints yielded consistently better models than the rest of the datasets (in 1,535 out of 2,764 situations). Results are displayed in Figure \ref{fig:fig_baseQSARperf_repres}. 

\begin{figure}
\center
\includegraphics[width=1.0\textwidth]{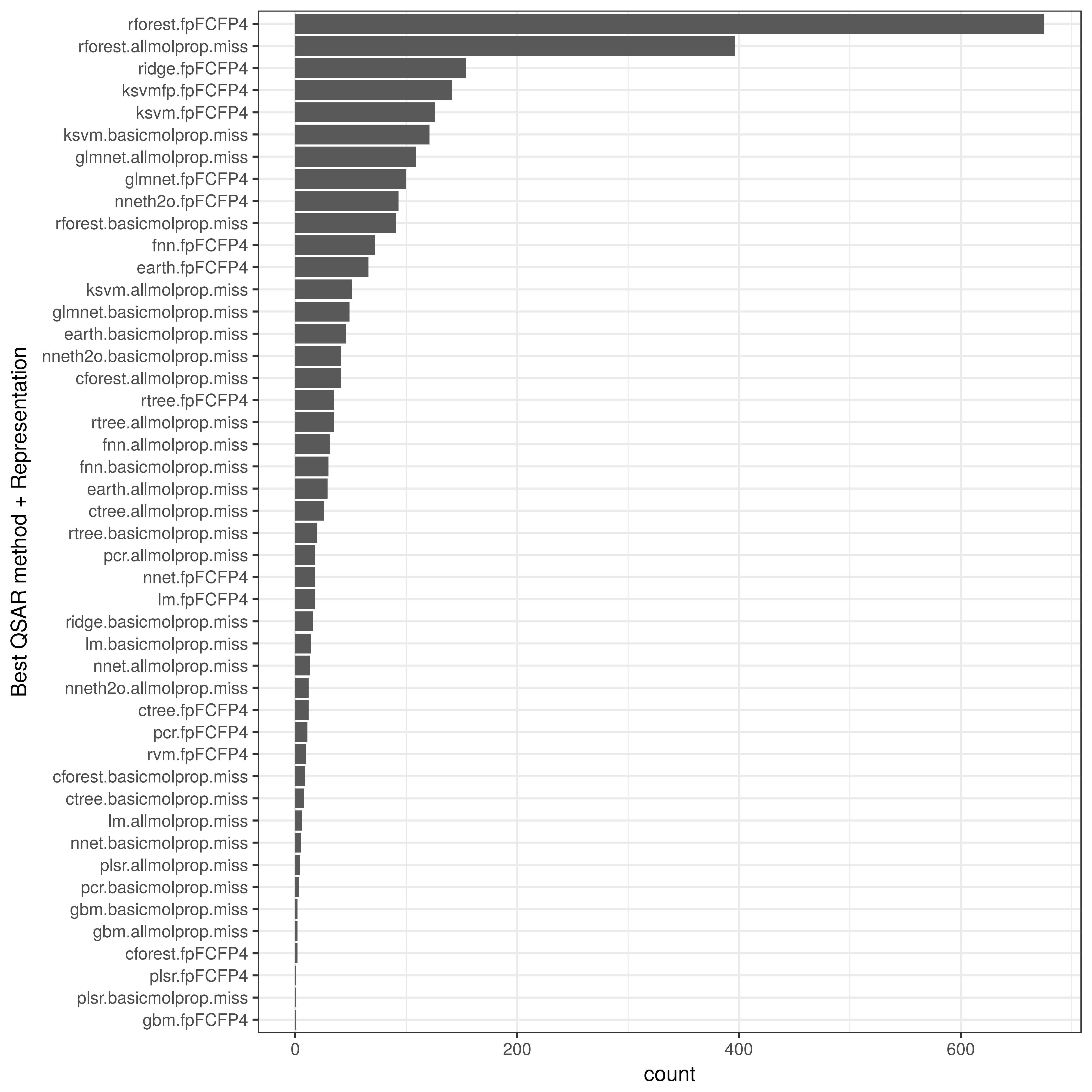}
\caption{Graphical representation of the number of times (target counts) a combination of dataset representation and QSAR method obtained the best performance (minimum RMSE)}
\label{fig:fig_baseQSARperf_strategies}
\end{figure}

Figure \ref{fig:fig_baseQSARperf_strategies} summarizes the results obtained using various strategies (combinations of QSAR algorithm and dataset representation). As the figure shows, the bar plot is highly skewed towards the top ranked QSAR strategies with a long tail representing QSAR problems in which other algorithms perform better. 

Applying random forest to datasets formed using either FCFP4 fingerprints or all molecular properties were the most successful QSAR strategies (in the figure, rforest.fpFCFP4 for 675 and rforest.allmolprop.miss for 396 out of 2,764 targets, respectively). Other strategies, such as regression with ridge penalisation (ridge.fpFCFP4), SVM with Tanimoto kernel (ksvmfp.fpFCFP4), and SVM with RBF kernel (ksvm.fpFCFP4) were particularly successful when using the FCFP4 fingerprint dataset representation (for 154, 141, and 126 targets, respectively). The full list of strategies ranked by frequency of success is shown in the figure. Combinations that never produced best performances are not shown.

\begin{figure}
\center
\includegraphics[width=1.0\textwidth]{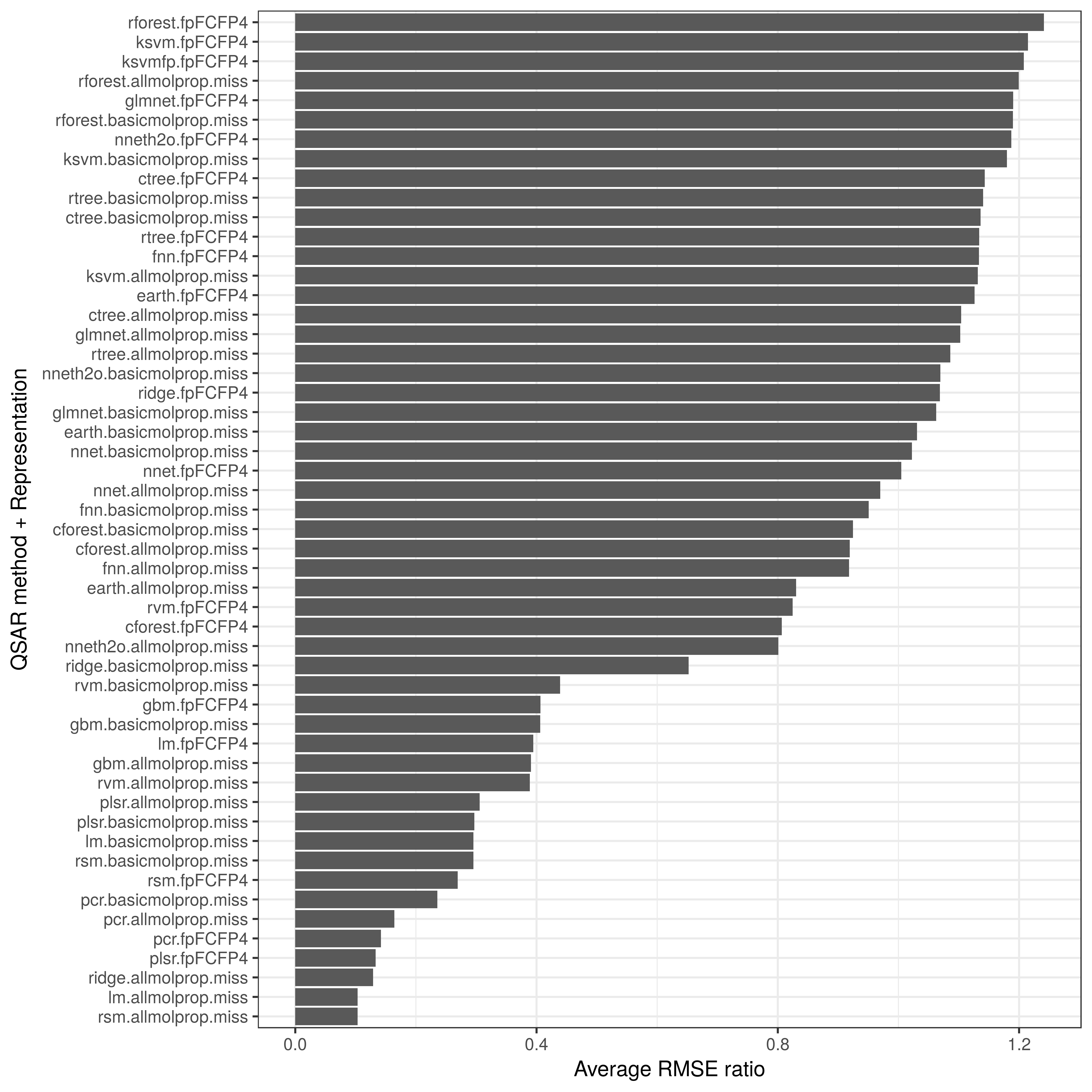}
\caption{Average ranking of dataset representation and QSAR combination as estimated using the RMSE ratio.}
\label{fig:fig_baseQSARperf_ranking}
\end{figure}

Combinations of QSARs and representations were also ranked by their average performances. For this, we estimated an average RMSE ratio score (aRMSEr) which is adapted from \cite{Brazdil:2003p697}, originally introduced for classification tasks. Our score was formulated as follows:

$$aRMSEr_{p}=\frac{\sum_{q}\sqrt[n]{\prod_{i}RMSEr_{p,q}^{i}}}{m}$$

where $RMSEr_{p,q}^{i}=RMSE_{q}^{i}/RMSE_{p}^{i}$ is the (inverse) RMSE ratio between algorithms $p$ and $q$ for the dataset $i$. In the same equation, $m$ represents the number of algorithms, whilst $n$, the number of targets. Notice that, an $RMSEr_{p,q}^{i}>1$ indicates that algorithm $p$ outperformed algorithm $q$. Ranking results using aRMSEr are presented in Figure \ref{fig:fig_baseQSARperf_ranking}.

We ran a Friedman test with a corresponding pairwise post-hoc test \cite{Demsar:2006p14241}, which is a non-parametric equivalent of ANOVA in order to verify whether the performances of baseline QSAR strategies were statistically different. The Friedman test ranks the strategies used per dataset according to their performance and tests them against the null hypothesis that they are equivalent. A post-hoc test was carried out if the null hypothesis is rejected. For this we used the Nemenyi test, also suggested by \citet{Demsar:2006p14241}. The resulting p-value ($10E-06$) from the test indicates the null hypothesis was invalid (p-value $<< 0.05$), which suggests that algorithm selection should significantly impact the overall performance.

\begin{figure}
\center
\includegraphics[width=1.0\textwidth]{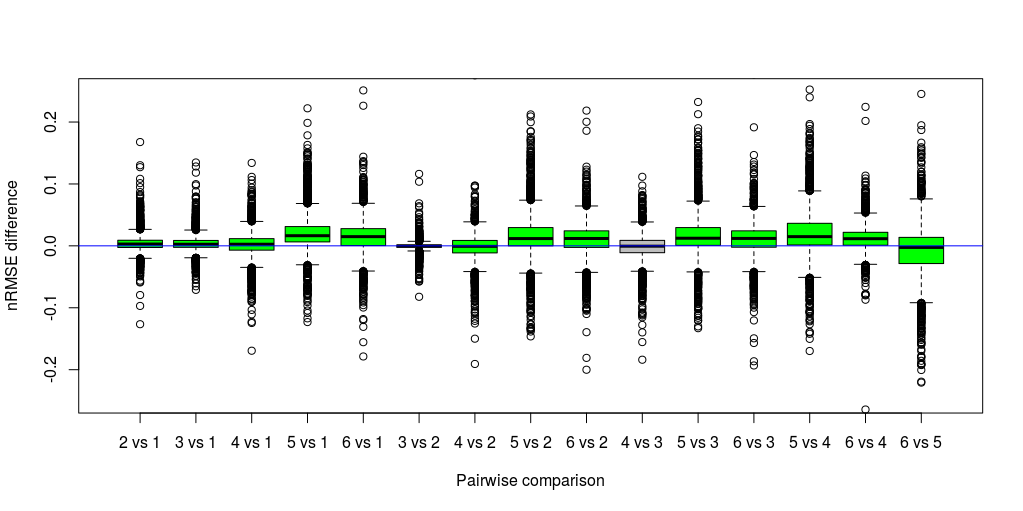}
\caption{Box plot displays the post-hoc test results over the top 6 ranked best performer QSAR strategies: 1 - rforest.fpFCFP4, 2 - ksvm.fpFCFP4, 3 - ksvmfp.fpFCFP4, 4 - rforest.allmolprop.miss, 5 - glmnet.fpFCFP4, and 6 - rforest.basicmolprop.miss.fs. Statistically significant comparisons (p-value $<0.05$) represented with green boxes.}
\label{fig:post-hoc-test}
\end{figure}

We ran the aforementioned post-hoc test for the top 6 QSAR strategies\footnote{Testing all possible pairwise combinations of QSAR strategies was not possible as the post-hoc test was running extremely slowly and we considered it would not add to the analyses of the results.} presented in Figure \ref{fig:fig_baseQSARperf_ranking}. Results are shown in Figure \ref{fig:post-hoc-test}. It shows that performance differences between the QSAR strategies were statistically significant with the exception of rforest.allmolprop.miss vs ksvmfp.fpFCFP4.

\section{Meta-features for meta-QSAR learning}
\label{sec:metafeatures}


\subsection{Meta-QSAR ontology}
\label{meta-qsar-ontology}

 Meta-learning analysis requires a set of meta-features. In our meta-qsar study we used measurable characteristics of the considered in the base study datasets and drug target properties as meta-features.
  We utilised a similar approach employed by BODO (the Blue Obelisk Descriptor Ontology) \cite{BODO} and the Chemical Information Ontology \cite{ChemInf}
for the formal definitions of molecular descriptors used in QSAR studies, and developed a   
  meta-qsar ontology\footnote{The ontology is available at \url{ https://github.com/larisa-soldatova/meta-qsar}}.
  
\begin{figure}
\center
\includegraphics[width=0.9\textwidth]{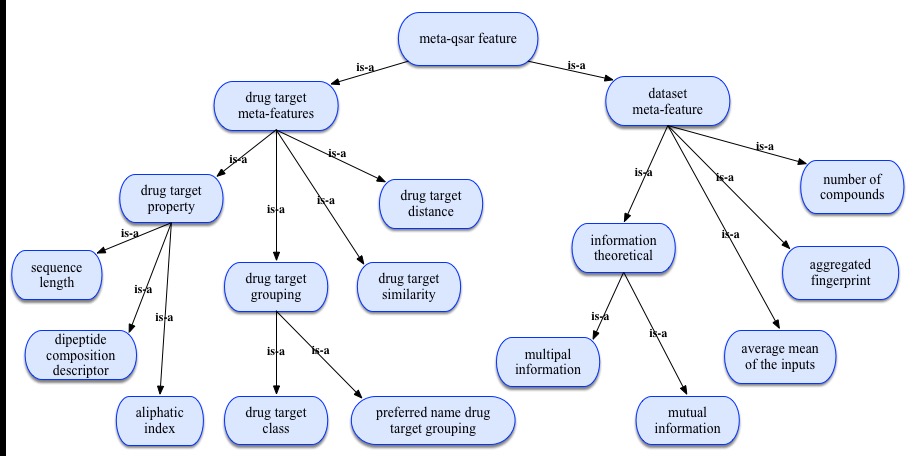}
\caption{The key branches of the meta-qsar ontology (a fragment). }
\label{fig:meta-qsar-ontology}
\end{figure}

\begin{figure}
\center
\includegraphics[width=0.7\textwidth]{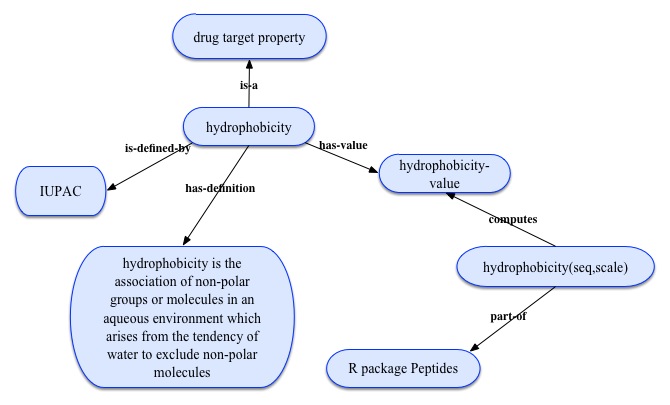}
\caption{The representation of the meta-features and their values.}
\label{fig:rep-of-meta-qsar-features}
\end{figure}

  The meta-qsar ontology provides formal definitions for the meta-features used in the reported meta-qsar study (see Figure \ref{fig:meta-qsar-ontology}). The meta-features are defined at the conceptual level, meaning that the ontology does not contain instance-level values of meta-features for each of 16,584 considered dataset. For example, the meta-feature 'multiple information' is defined as the meta-feature of a dataset (multiple information (also called total correlation) among the random variables in the dataset), but the meta-qsar ontology does not contain values of this meta-feature for each dataset. Instead, it contains  links to the code to calculate values of the relevant features. For example, we used the R Package Peptides\footnote{\url{http://cran.r-project.org/web/packages/Peptides/}} to calculate values of the meta-feature `hydrophobicity'. Figure \ref{fig:rep-of-meta-qsar-features} shows how this information is captured in the meta-qsar ontology. The description of the selected meta-features and instructions on the calculation of their values are available online\footnote{\url{https://github.com/meta-QSAR/drug-target-descriptors}}.

\subsection{Dataset meta-features}
\label{dataset-meta-features}
The considered 16,584 datasets have a range of different properties, e.g. 'number of compounds' (instances) in the dataset, 'entropy' and 'skewness' of the features and  'target meta-feature', 'mutual information' and 'total correlation' between the input and output features (see Table \ref{table:dataset-meta-features} for more detail). The dataset properties have a significant effect on the performance of the explored algorithms and were used for the meta-qsar learning. 
Figure \ref{fig:fig_variable_importance} shows the level of influence of different categories of meta-features. For example information-theoretical meta-features make a considerable contribution to  meta-learning.

\begin{table}
\begin{tabularx}{\columnwidth}{ | l | >{\raggedright\arraybackslash} X | }
\hline
Feature & Description \\ \hline
multiinfo &  Multiple information  (also called total correlation) among the random variables in the dataset. \\ \hline
mutualinfo &  Mutual information between nominal attributes X and Y. Describes the reduction in uncertainty of Y due to the knowledge of X, and leans on the conditional entropy $H(Y|X)$.\\ \hline
nentropyfeat & Normalised entropy of the features which is the class entropy divided by log(n) where n is the number of the features.\\ \hline
mmeanfeat &  Average mean of the features.\\ \hline
msdfeat &  Average standard deviation of the features.\\ \hline
kurtresp &  Kurtosis of the response variable.\\ \hline
meanresp &  Mean of the response variable.\\ \hline
skewresp &  Skewness of the response variable.\\ \hline
nentropyresp &  Normalised entropy of the response variable.\\ \hline
sdresp & Standard deviation of the response.\\ \hline
aggFCFP4fp (1024 features) & Aggregated fingerprints and normalized over the number of instances in the dataset.\\ \hline
\end{tabularx}
\caption{Dataset meta-features (Examples).}
\label{table:dataset-meta-features}
\end{table}

Some descriptors of the dataset properties, e.g. 'number of instances', have been imported from the Data Mining Optimization (DMOP) Ontology \footnote{\url{www.dmo-foundry.org/DMOP}} \cite{DMOP}. We also added qsar-specific dataset descriptors 'aggregated fingerprint'. These were calculated by summing 1s (set bits) in each of the 1024 columns and normalised by the number of the compounds in each dataset.

\subsection{Drug target meta-features}
\label{drug-target-metafeatures}

\subsubsection{Drug target properties}
\label{drug-target-props}
The QSAR datasets are additionally characterized by measurable properties of the drug target (a protein) they represent, such as 'aliphatic index', 'sequence length', 'isoelectric point' (see Table \ref{table:drug-target-meta-features} for more details). These differ from the molecular properties we used to describe the chemical compounds in the QSAR dataset instances, e.g. 'molecular weight' (MW), 'LogP'.


\begin{table}[h]
\centering
\begin{tabularx}{\columnwidth}{ | l | >{\raggedright\arraybackslash} X | }
\hline
Feature & Description \\ \hline
Aliphatic index &
The Aliphatic index~\cite{IKAI1980} is defined as the relative volume occupied by aliphatic side chains (Alanine, Valine, Isoleucine, and Leucine). It may be regarded as a positive factor for the increase of thermo stability of globular proteins.\\ \hline
Hydrophobicity & Hydrophobicity is the association of non-polar groups or molecules in an aqueous environment which arises from the tendency of water to exclude non-polar molecules \cite{IUPAC}.\\ \hline
Boman index & This the potential protein interaction index proposed by Boman~\cite{boman2003}. It is calculated as the sum of the solubility values for all residues in a sequence~\cite{peptides}. \\ \hline
Hydrophobicity (38 features) & Hydrophobicity is the association of non-polar groups or molecules in an aqueous environment which arises from the tendency of water to exclude non-polar molecules \cite{IUPAC}. We estimated 38 variants of hydrophobicity.\\ \hline
Net charge & The theoretical net charge of a protein sequence as described by Moore~\cite{moore1985}.\\ \hline
Molecular weight & Ratio of the mass of a molecule to the unified atomic mass unit. Sometimes called the molecular weight or relative molar mass \cite{IUPAC}.\\ \hline
Isoelectric point & The pH value at which the net electric charge of an elementary entity is zero. (pI is a commonly used symbol for this kind-of-quantity, however more accurate symbol is pH(I)) \cite{IUPAC}. \\ \hline
Sequence length & A number of amino acids in a protein sequence. \\ \hline
Instability index & The instability index was proposed by (Guruprasad, 1990). A protein whose instability index is smaller than 40 is predicted as stable, a value above 40 predicts that the protein may be unstable.\\ \hline
DC groups (400 features) & The Dipeptide Composition descriptor~\cite{protr2015,bhasin2004} captures information about the fraction and local order of amino acids.\\ \hline
\end{tabularx}
\caption{Drug targets meta-features (Examples).}
\label{table:drug-target-meta-features}
\end{table}

\subsubsection{Drug target groupings}
\label{drug-target-groupings}
We also used drug target groupings~\cite{citeulike:881883}, such as 'drug target classes', and 'the preferred name groupings', as meta-features.
These enable meta-learning to exploit known biological/chemical relationships between the targets (proteins). Indeed, if the target proteins are similar, this may make the resulting datasets more similar too.
\\
\textbf{Drug target classes:}
The ChEMBL database curators have classified the protein targets in a manually curated family hierarchy. The version of the hierarchy that we have used (taken from ChEMBL20) comprises 6 levels, with Level 1 (L1) being the broadest class, and Level 6 (L6) the most specific. For example, the protein target `Tyrosine-protein kinase Srms' is classified as follows: Enzyme (L1), Kinase (L2), Protein Kinase (L3), TK protein kinase group (L4), Tyrosine protein kinase Src family (L5), Tyrosine protein kinase Srm (L6). Different classes in Level 1 are not evolutionarily related to one another, whereas members of classes in L3 and below generally share common evolutionary origins. The picture is mixed for L2. The hierarchy is not fully populated, with the greatest emphasis being placed on the target families of highest pharmaceutical interest, and the different levels of the hierarchy are not defined by rigorous criteria. However, the hierarchical classification provides a useful means of grouping related targets at different levels of granularity.
                                                        	
\noindent \textbf{The preferred name drug targets grouping:}
The ChEMBL curators have also assigned each protein target a \emph{preferred name} - in a robust and consistent manner, independent of the various adopted names and synonyms used elsewhere. This preferred name is based on the practice that individual proteins can be described by a range of different identifiers and textual descriptions across the various data resources. The detailed manual annotation of canonical target names means that, for the most part, orthologous proteins (evolutionarily related proteins with the same function) from related species are described consistently, allowing the most related proteins to be grouped together. In the preferred name groupings, we obtained 468 drug target groups, each with two or more drug targets. The largest drug target group is that of Dihydrofolate Reductase with 21 drug targets.

\section{Meta-Learning: QSAR Algorithm Selection}
\label{sec:meta-algo}
\begin{figure}
\center
\includegraphics[width=0.9\textwidth]{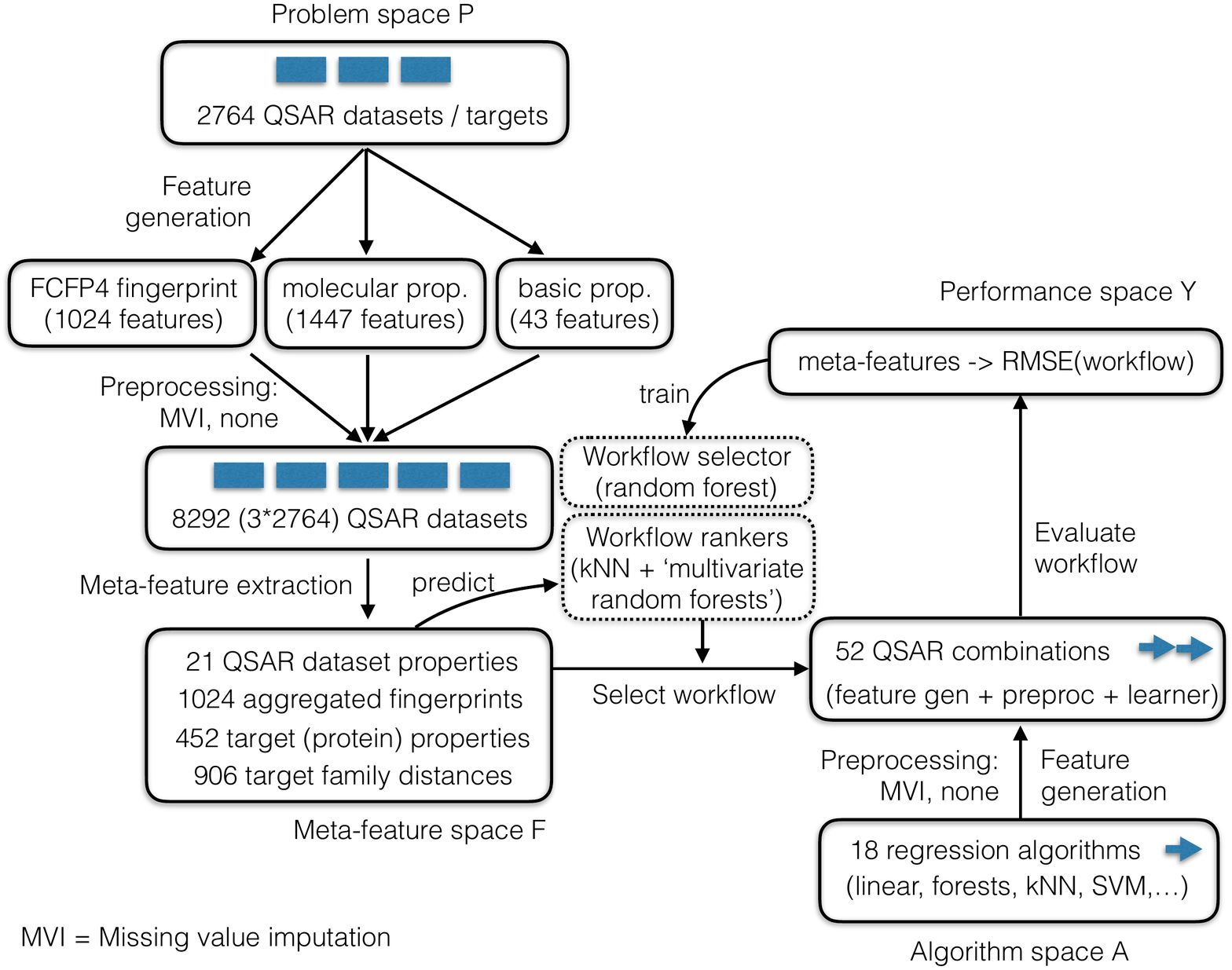}
\caption{Meta-learning setup to select QSAR combinations (workflows) for a given QSAR dataset. The 52 QSAR combinations are generated by combining 3 types of representation/preprocessing with 17 regression algorithms, plus the Tanimoto KSVM which was only run on the fingerprint representation.}
\label{fig:qsarframework}
\end{figure}

 We cast the meta-QSAR problem as two different problems: 1) the classification task to predict which QSAR method should be used for a particular QSAR problem; and 2) ranking prediction task to rank QSAR methods by their performances. This entails a number of extensions to Rice's framework in Figure \ref{rice}, as we are now dealing with multiple dataset representations per QSAR problem, and learning algorithm. The resulting setup is shown in Figure \ref{fig:qsarframework}. Each original QSAR problem is first represented in 3 different ways resulting in 3 datasets for each QSAR target, from which we extract 11 dataset-based meta-features each (see Section \ref{dataset-meta-features})\footnote{The actual number (21) is slightly smaller because some meta-features, such as the number of instances, is identical for each dataset.}, as well as over 450 meta-features based on the target (protein) that the dataset represents (see Section \ref{drug-target-metafeatures}). The space of algorithms consists of workflows that generate the base-level features, and run one of the the 18 regression algorithms (see Section \ref{baseline-algorithms}), resulting in 52 workflows which are evaluated, based on their RMSE, on the corresponding datasets (those with the same representation).

\subsection{Meta-QSAR dataset}
\label{meta-learning-dataset}
\begin{figure}
\center
\includegraphics[width=1.0\textwidth]{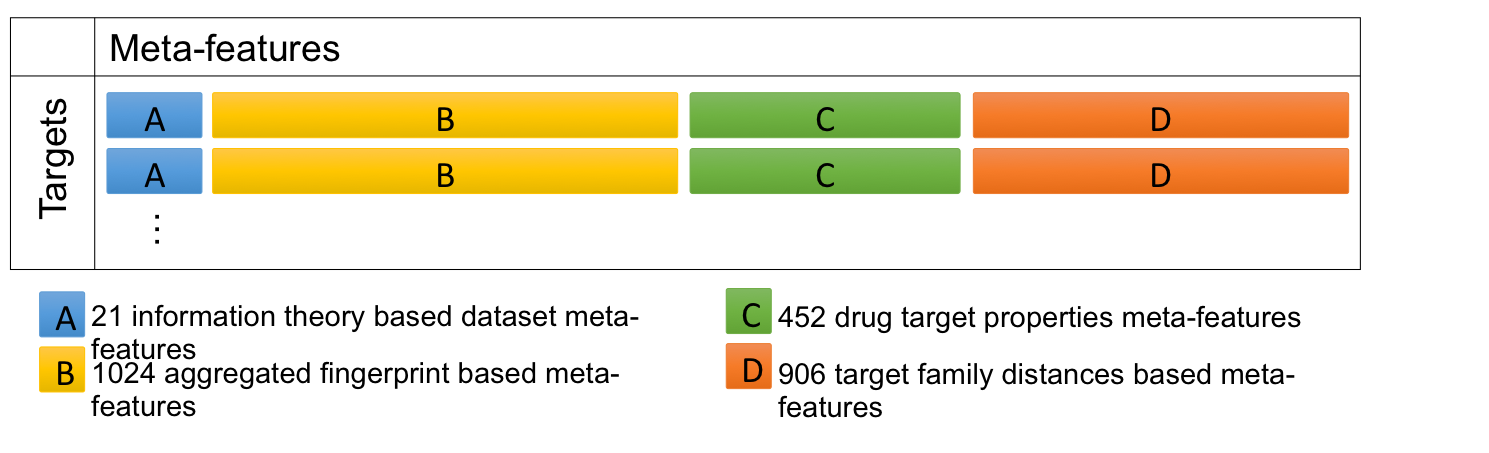}
\caption{Schematic representation of the meta-dataset used for meta-QSAR.}
\label{fig:fig_schem_meta-dataset}
\end{figure}

A training meta-dataset was formed using the meta-features extracted from the baseline QSAR datasets as the inputs. For the classification tasks we used the best QSAR strategy (combination of QSAR method and dataset representation) per target as the output labels, whilst for the ranking tasks, the QSAR performances (RMSEs) were used. Figure \ref{fig:fig_schem_meta-dataset} shows a schematic representation of the meta-dataset used in the meta-learning experiments. As this figure shows, we used meta-features derived from dataset and drug target properties. The size of the final meta-dataset was 2,394 meta-features by 2,764 targets.

\subsection{Meta-QSAR learning algorithms}
\label{meta-learning-methods}

A meta-learning classification problem using all possible combinations of QSAR methods and dataset representions was implemented using a random forest with 500 trees. Given the large number of classes (52 combinations) and the highly imbalanced classification problem (as shown in Figure \ref{fig:fig_baseQSARperf_strategies}, additional random forest implementations using the top 2, 3, 6, 11 and 16 combinations (Figure \ref{fig:fig_baseQSARperf_ranking}) were also investigated. For the ranking problem, we used two approaches: K-nearest neighbour approach (k-NN), as suggested in \cite{Brazdil:2003p697}, and a multi-target regression approach. Experiments with k-NN were carried out using 1, 5, 10, 50, 100, 500, and all neighbours. The multi-target regression was implemented using a multivariate random forest regression \cite{Segal2011} with 500 trees to predict QSAR performances and with them, to rank QSAR combinations. All implementations were assessed using 10-fold cross-validation.

\subsection{Results}
\label{meta-qsar-results}

\begin{figure}
\center
\includegraphics[width=0.8\textwidth]{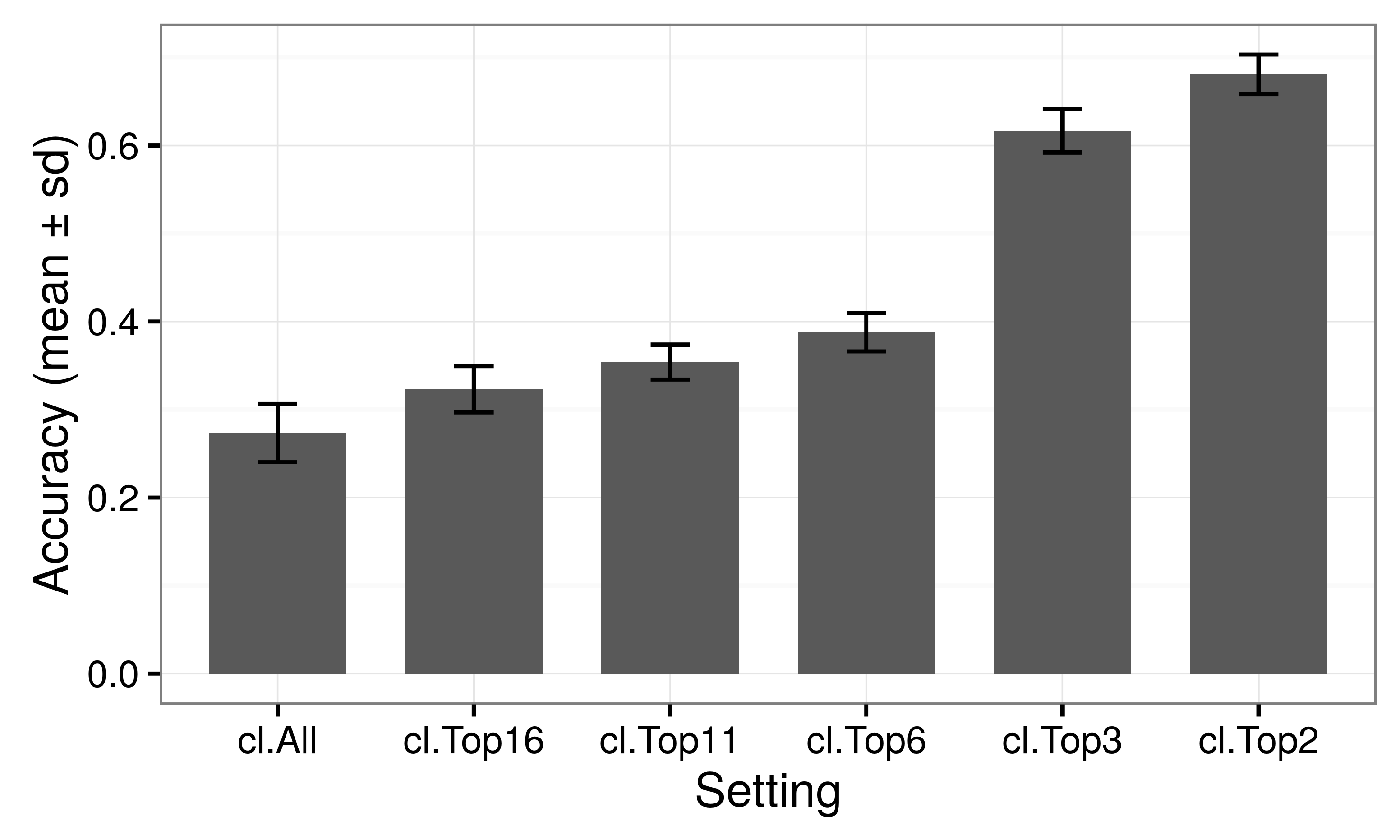}
\caption{Accuracies for 6 classification settings considered for meta-learning: cl.All (52 QSAR combinations), cl.Top16, cl.Top11, cl.Top6, cl.Top3, and cl.Top2 (best 16, 11, 6, 3, and 2 ranked QSAR combinations according to Figure \ref{fig:fig_baseQSARperf_ranking})}
\label{fig:fig_accuracies_classif}
\end{figure}

Algorithm selection experiments were applied to the 6 classification problems defined above. Results of the classification performances are presented in Figure \ref{fig:fig_accuracies_classif} in the form of classification accuracies. As can be observed in the figure, performances improve as the number of base-learners decreases.

\begin{figure}
\center
\includegraphics[width=1.0\textwidth]{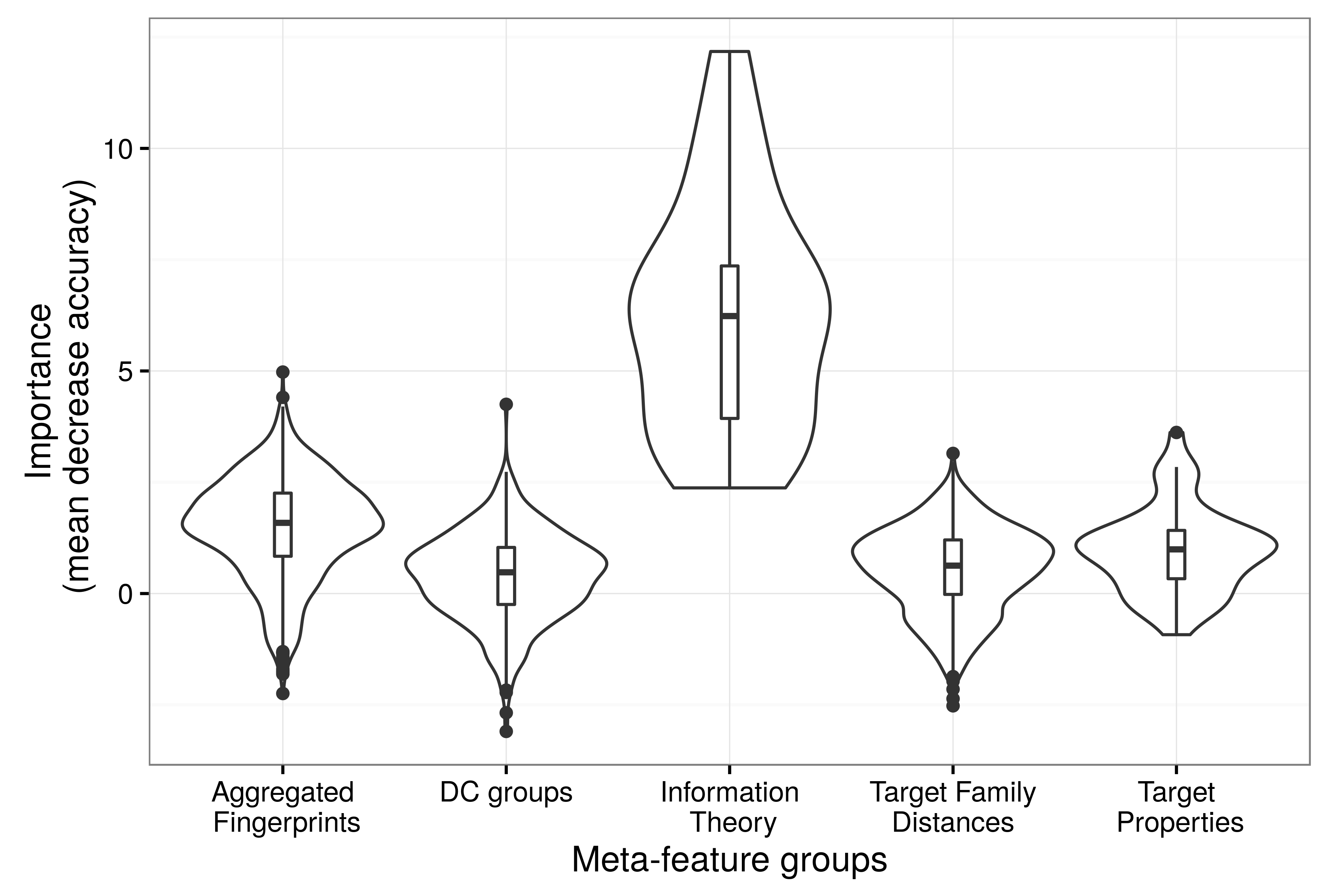}
\caption{Violin plots with added box plots representing the mean decrease accuracy of the meta-features grouped by meta-feature groups. Notice that for visualization purpose, we are showing the group dataset meta-features (as defined in Section 3) in two separated groups: "Aggregated Fingerprints" and "Information Theory".}
\label{fig:fig_variable_importance}
\end{figure}

We also use the all-classes random forest implementation to estimate the importance of each meta-feature in the classification task, as estimated using the mean decrease accuracy. Summary results considered by meta-feature groups are presented in Figure \ref{fig:fig_variable_importance}. It is seen that the meta-features belonging to the information theory group (all dataset meta-features but the aggregated fingerprints, Table \ref{table:dataset-meta-features}) were the most relevant, although we found all groups contributed to the task. 

\begin{figure}
\center
\includegraphics[width=0.9\textwidth]{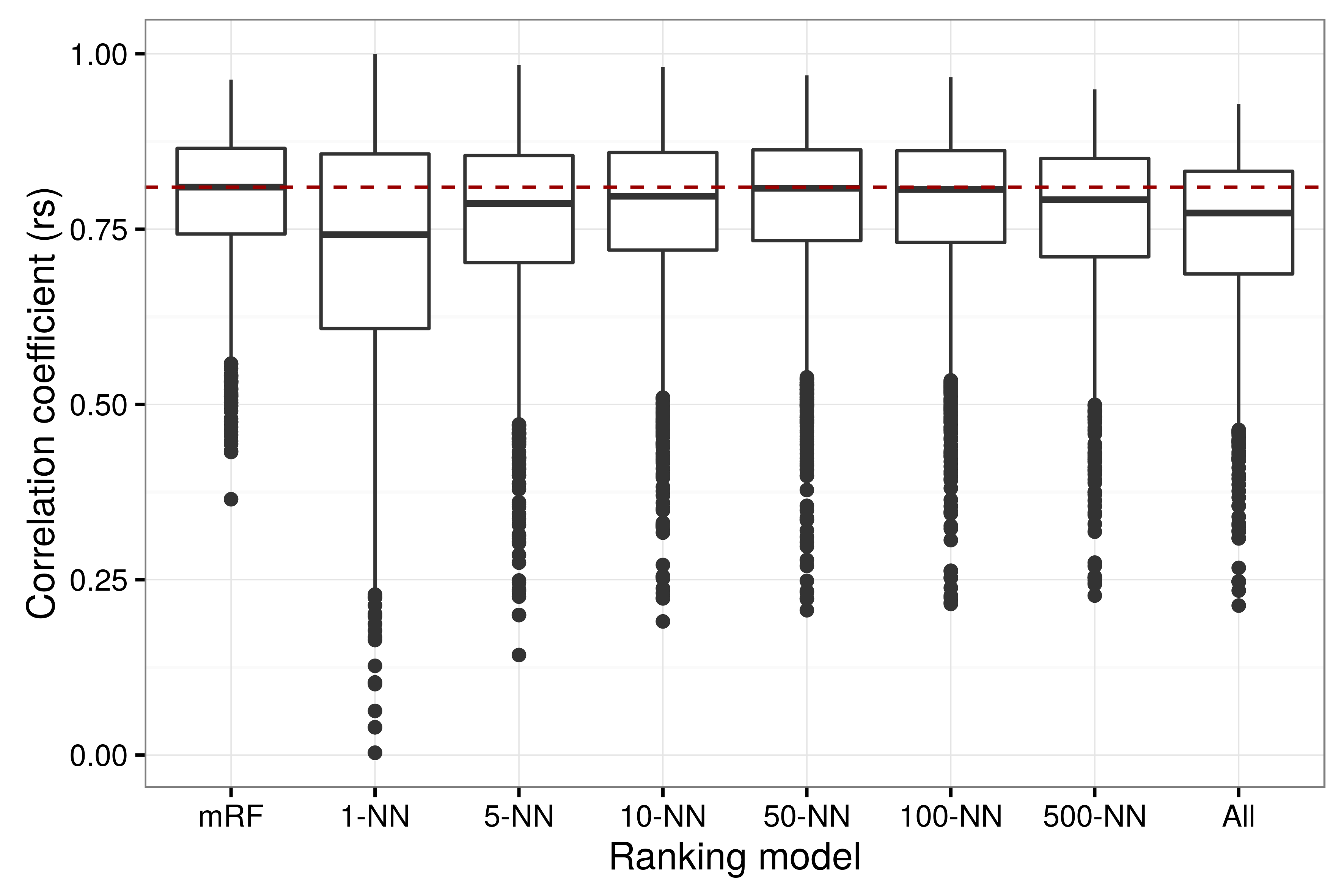}
\caption{Box plots representing the computed Spearman's rank correlation coefficient (rs) between the predicted and actual rankings. Labels in the horizontal axis indicates: mRF - multivariate random forest, 1-NN, 5-NN, 10-NN, 50-NN, 100-NN, 500-NN, and All - 1, 5, 10, 50, 100, 500, and all nearest neighbours, respectively.} 
\label{fig:fig_spearman_correlation_boxplots}
\end{figure}

\begin{figure}
\center
\includegraphics[width=1.0\textwidth]{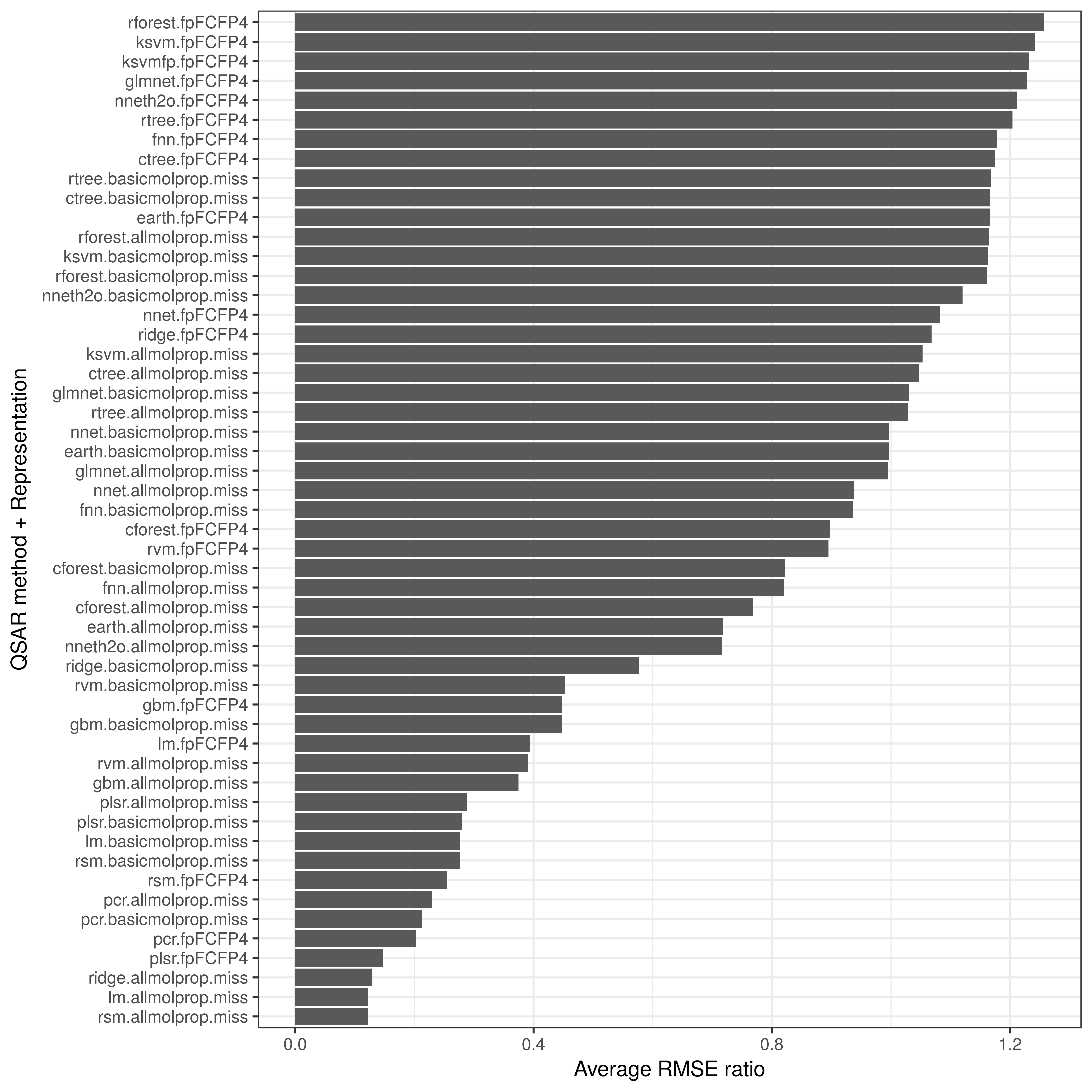}
\caption{Average of predicted ranking of QSAR combinations using the multivariate random forest algorithm according to the RMSE ratio.} 
\label{fig:fig_predicted_ranking}
\end{figure}

As mentioned before, k-NN and multivariate random forest were used to implement ranking models. We used the Spearman's rank correlation coefficient to compare the predicted with the actual rankings (average of the actual rankings were shown in Figure \ref{fig:fig_baseQSARperf_ranking}). Results of these comparisons are shown in Figure \ref{fig:fig_spearman_correlation_boxplots}. It is observed from the figure that the multivariate random forest and 50-nearest neighbours implementations (mRF and 50-NN in the figure) predicted better rankings, overall. For illustrative purpose, the average of the predicted rankings by multivariate random forest is displayed in Figure \ref{fig:fig_predicted_ranking}.

\begin{figure}
\center
\includegraphics[width=1.0\textwidth]{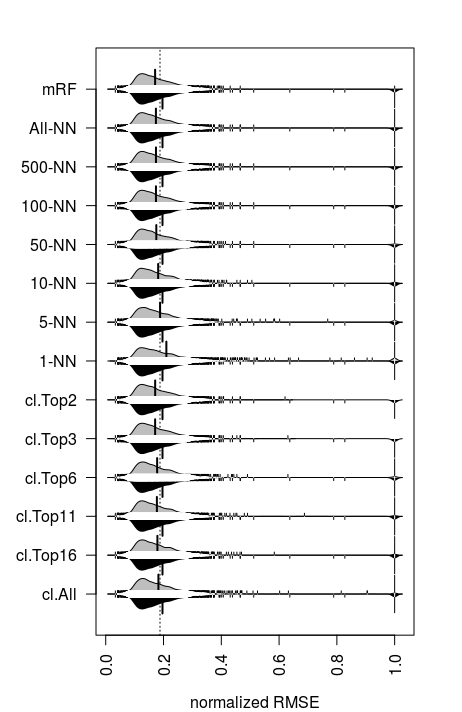}
\caption{Visual comparison of performance distributions between the default strategy (in black) and all meta-learners (in grey) using asymmetric bean plots. Average RMSE for each implementation is represented by vertical black lines on the "beans" (performance distribution curves). }
\label{fig:fig_metaqsar_perf_comparison}
\end{figure}

Performances of the best suggested QSAR combination by all Meta-QSAR implementations were compared with an assumed default. In the case of the ranking models, the best suggested QSAR combination is the one ranked the highest in each QSAR problem. For the default (baseline) we used random forest with the fingerprint molecular representation (“rforest.fpFCFP4”), as this is well-known for its robust and reliable performance (Fig. \ref{fig:fig_baseQSARperf_strategies}), and hence represents a strong baseline. Results are shown in Figure \ref{fig:fig_metaqsar_perf_comparison}. As it can be observed in this figure, most of the Meta-QSAR implementations improved overall performance in comparison with the default QSAR combination with the exception of the 1-nearest neighbour. These results suggest that meta-learning can be successfully used to select QSAR algorithm/representation pairs that perform  better than the best algorithm/representation pair (default strategy).

\section{Discussion}
\label{sec:discussion}
QSARs models are regression models, empirical functions that relate a quantitative description of a chemical structure (a drug) to some form of biological activity (e.g. inhibiting proteins) for the purposes of informing drug design decision-making. Many consider the seminal papers of Hansch et al. \cite{Hansch:1964ea} to be the origin of the QSAR field. Since then, such predictive modelling approaches have grown to become a core part of the drug discovery process \cite{Cumming:2013eu,Cherkasov:2014go}.
The subject is still increasing in importance \cite{Cramer:2012bk}. This  may be attributed to the alignment of a number of factors, including improving availability of data, advances in data-mining methodologies as well as a more widespread appreciation of how to avoid many of numerous pitfalls in building and applying QSAR models \cite{Cherkasov:2014go}. Current trends in the field include efforts in chemical data curation \cite{Williams:2012bo}, automation of QSAR model building \cite{Cox:2013dq}, exploration of alternative descriptors \cite{Cherkasov:2014go}, and efforts to help define the Applicability Domain (AD) of a given QSAR model \cite{Sahigara:2012kb}.

To facilitate application of QSAR models in the drug regulatory process, the Organization for Economic Co-operation and Development (OECD) has provided guidance to encourage good practice in QSAR modelling. The OECD guidelines recommend that a QSAR model has i) a defined end point; ii) an unambiguous algorithm; iii) a defined domain of applicability; iv) appropriate measures of goodness of fit, robustness and predictivity; and v) a mechanistic interpretation, if possible. However, the application of QSAR models in drug discovery is still fraught with difficulties, not least because the model builder is faced with myriad options with respect to choice of descriptors and machine learning methods. 

The application of meta-learning in this study helps ameliorate this issue by providing some guidance as to which individual method performs the best overall as well as which method may be the most appropriate given the particular circumstances.    

Our comparison of QSAR learning methods involves 18 regression methods and 3 molecular representations applied to more than 2,700 QSAR problems, making it one of the most extensive ever comparisons of base learning methods reported. Moreover, the QSAR datasets, source code, and all our experiments are available on OpenML~\cite{Vanschoren:2013:2641190.2641198} \footnote{See \url{http://www.openml.org/s/13}}, so that our results can be easily reproduced. This is not only a valuable resource for further work in drug discovery, it will foster the development of meta-learning methods as well. Indeed, as all the experimental details are fully available, there is no need to run the baseline-learners again, so research effort can be focused on developing novel meta-learning methods.

In this paper we have investigated algorithm selection for QSAR learning.
Note however, that many more meta-learning approaches could be applied: it would be interesting to investigate other algorithm selection methods (see Section \ref{previous-work}), such as other algorithm ranking approaches (e.g. active testing or collaborative filtering), and model-based optimization. Another alternative framing of the meta-learning problem would be to use a regression algorithm on the meta-level and predict the \textit{performance} of various regression algorithms. We will explore this in future work. Finally, we would also like to explore other algorithms selection techniques beyond Random Forests. To this end, we plan to export our experiments from OpenML to an ASlib scenario \cite{aslib}, where many algorithm selection techniques could be compared.

The success of meta-learning crucially depends on having a large set of datasets to train a meta-learning algorithm, or simply to find similar prior datasets from which best solutions could be retrieved. This work provides more than 16,000 datasets, which is several orders of magnitude larger than what was available before. It has often been observed that machine learning breakthroughs are being made by having novel large collections of data: ImageNet\footnote{http://www.image-net.org}, for instance, sparked breakthroughs in image recognition with deep learning. The datasets made available here could have a similar effect in accelerating meta-learning research, as well as novel machine learning solutions for drug discovery. Moreover, it is but the first example of what is possible if large collections of scientific data are made available as readily usable datasets for machine learning research. Beyond ChEMBL, there exist many more databases in the life sciences and other fields (e.g. physics and astronomy), which face similar challenges in selecting the best learning algorithms, hence opening up interesting further avenues for meta-learning research.

Beyond the number of datasets, this study pushes meta-learning research in several other ways. First, it is one of the few recent studies focussing on regression problems rather than classification problems. Second, it uses several thousands (often domain-specific) meta-features, which is much larger than most other reported studies. And third, it considers not only single learning algorithms, but also (small) workflows consisting of both preprocessing and learning algorithms.

There is ample opportunity for future work. For instance, besides recommending the best algorithm, one could recommend the best hyperparameter settings as well (e.g. using model-based optimization). Moreover, we did not yet include several types of meta-features, such as landmarkers or model-based meta-features, which could further improve performance. Finally, instead of using a RandomForest meta-learner, other algorithms could be tried as well. One particularly interesting approach would be to use Stacking \cite{Wolpert:1992p4366} to combine all the individually learned models into a larger model that exploits the varying quantitative predictions of the different base-learner and molecular representation combinations. 
However developing such a system is more computationally complex than simple algorithm selection, as it requires applying cross-validation over the base learners.

\section{Conclusions}
\label{sec:conclusions}

QSAR learning is one of the most important and established applications of machine learning. We demonstrate that meta-learning can be leveraged to build QSAR models which are much better than those learned with any base-level regression algorithm. We carried out the most comprehensive ever comparison of machine learning methods for QSAR learning: 18 regression methods, 6 molecular representations, applied to more than 2,700 QSAR problems. This enabled us to first compare the success of different base-learning methods, and then to compare these results with meta-learning. We found that algorithm selection significantly outperforms the best individual QSAR learning method (random forests using a molecular fingerprint representation). The application of meta-learning in this study helps accelerate research in drug discovery by providing guidance as to which machine learning method may be the most appropriate given particular circumstances. Moreover, it represents one of the most extensive meta-learning studies ever, including over 16,000 datasets and several thousands of meat-features. The success of meta-learning in QSAR learning provides evidence for the general effectiveness of meta-learning over base-learning, and opens up novel avenues for large-scale meta-learning research.

\section*{Acknowledgements}
This research was funded by the Engineering and Physical Sciences Research Council (EPSRC) grant EP/K030469/1.

\bibliographystyle{plainnat}
\bibliography{references}
\end{document}